\documentclass[letterpaper, 10pt, conference]{ieeeconf}
\IEEEoverridecommandlockouts

\usepackage{cite}
\usepackage{amsmath,amssymb,amsfonts}
\usepackage{algorithmic}
\usepackage{graphicx}
\usepackage{textcomp}
\def\BibTeX{{\rm B\kern-.05em{\sc i\kern-.025em b}\kern-.08em
    T\kern-.1667em\lower.7ex\hbox{E}\kern-.125emX}}
\usepackage{euscript}
\usepackage{xcolor}
\usepackage{epsfig}

\usepackage{mathtools}
\mathtoolsset{showonlyrefs}

\usepackage{listings}
\usepackage{booktabs} 
\usepackage{longtable}
\usepackage{multirow}
\usepackage{graphicx}
\usepackage{epstopdf}
\usepackage{bm}
\usepackage{makecell}
\usepackage{subcaption}
\usepackage{matlab-prettifier}
\usepackage{mathrsfs}
\usepackage{tikz}
\usetikzlibrary{shapes,arrows}
\usepackage{verbatim}
\usepackage[normalem]{ulem} 

\tikzstyle{block} = [draw, rectangle, 
    minimum height=3em, minimum width=6em]
\tikzstyle{sum} = [draw, circle, node distance=1cm]
\tikzstyle{input} = [coordinate]
\tikzstyle{output} = [coordinate]
\tikzstyle{pinstyle} = [pin edge={to-,thin,black}]

\newtheorem{theorem}{\bf \text{Theorem}}

\newtheorem{assumption}{\bf \text{Assumption}}

\newtheorem{lemma}{\bf \text{Lemma}}
\newtheorem{remark}{\bf \text{Remark}}

\newcommand{\E}{{\mathbb E}}
\newcommand{\COV}{{\mathbb C\mathbb O\mathbb V}}

\newcommand{\R}{{\mathbb R}}
\newcommand{\C}{{\mathbb C}}

\newcommand{\TR}{\text{R}}

\makeatletter

\newcommand{\Rmnum}[1]{\expandafter \@slowromancap\romannumeral #1@}
\makeatother

\DeclareMathOperator*{\rank}{rank}

\DeclareMathOperator*{\Tr}{Tr}

\DeclareMathOperator*{\Var}{Var}

\DeclareMathOperator*{\EB}{EB}

\DeclareMathOperator*{\tb}{b}

\DeclareMathOperator*{\XMSE}{XMSE}
\DeclareMathOperator*{\ML}{ML}

\DeclareMathOperator*{\MSE}{MSE}

\DeclareMathOperator*{\XBias}{XBias}
\DeclareMathOperator*{\XVar}{XVar}
\DeclareMathOperator*{\XVarHPE}{XVarHPE}
\DeclareMathOperator*{\Bayes}{Bayes}

\DeclareMathOperator*{\Biased}{Biased}
\DeclareMathOperator*{\FIT}{FIT}

\DeclareMathOperator*\argmin{arg\,min}

\allowdisplaybreaks[4]

\newtheorem{subassumption}{\bf{Assumption\normalfont}}[assumption]

\newenvironment{assumption*}
 {\ifnum\value{subassumption}=0 \stepcounter{assumption}\fi\subassumption}
 {\endsubassumption}
\newenvironment{assumption+}[1]
 {\renewcommand{\thesubassumption}{#1} \subassumption}
 {\endsubassumption}

\begin{document}

\title{\LARGE\bf Bayes and Biased Estimators Without Hyper-parameter Estimation: Comparable Performance to the Empirical-Bayes-Based Regularized Estimator*
}

\author{Yue Ju$^{1}$, Bo Wahlberg$^{1}$, H\r{a}kan Hjalmarsson$^{1}$%
\thanks{*This work was supported by VINNOVA Competence Center AdBIOPRO, contract [2016-05181] and by the Swedish Research Council through the research environment NewLEADS (New Directions in Learning Dynamical Systems), contract [2016-06079], and contract [2019-04956].}%
\thanks{$^{1}$Yue Ju, Bo Wahlberg, and H\r{a}kan Hjalmarsson are now with Division of Decision and Control Systems, School of Electrical Engineering and Computer Science, KTH Royal Institute of Technology, 10044 Stockholm, Sweden (e-mails: yuej@kth.se, bo@kth.se, hjalmars@kth.se). H\r{a}kan Hjalmarsson is also with the Competence Centre for Advanced BioProduction by Continuous Processing, AdBIOPRO.}%
}

\maketitle
\thispagestyle{empty}
\pagestyle{empty}

\begin{abstract}
Regularized system identification has become a significant complement to more classical system identification. It has been numerically shown that kernel-based regularized estimators often perform better than the maximum likelihood estimator in terms of minimizing mean squared error (MSE). However, regularized estimators often require hyper-parameter estimation. This paper focuses on ridge regression and the regularized estimator by employing the empirical Bayes hyper-parameter estimator. We utilize the excess MSE to quantify the MSE difference between the empirical-Bayes-based regularized estimator and the maximum likelihood estimator for large sample sizes. We then exploit the excess MSE expressions to develop both a family of generalized Bayes estimators and a family of closed-form biased estimators. They have the same excess MSE as the empirical-Bayes-based regularized estimator but eliminate the need for hyper-parameter estimation. Moreover, we conduct numerical simulations to show that the performance of these new estimators is comparable to the empirical-Bayes-based regularized estimator, while computationally, they are more efficient.
\end{abstract}


\section{Introduction}\label{sec:introduction}
Classical system identification deals with the problem of estimating model parameters of dynamical systems based on observed data using the maximum likelihood (ML) or prediction error (PE) method \cite{Ljung:99}. For linear regression models with Gaussian distributed measurement noise with zero mean and known variance, it is well-known that the ML estimator is unbiased, see, e.g., \cite{Ljung:99, Greene2003}. However, the ML estimator may have a large variance and thus have a large MSE, in particular for an ill-conditioned regression matrix. In this case, a well-designed biased estimator can achieve a better bias-variance trade-off. 

Kernel-based regularized estimators, known as regularized system identification \cite{PCCDL2022, PDCDL2014}, are a family of biased estimators that complement classical system identification. The design, implementation, and theoretical analysis of regularized estimators have been thoroughly investigated. The design of kernels based on given information has been discussed in \cite{SS02, RasmussenW:06, RBH15, Chen18, KHC22, Z23, FC24}. {Efficient implementations} of regularized estimators and the corresponding hyper-parameter estimation have been considered in \cite{CL:12, VHW20, CA21, ZJMZC23, XFMC24}. The asymptotic properties of regularized estimators and their corresponding hyper-parameter estimators for large sample sizes have been explored in \cite{KF2000, UV2010, MCL18asy, JCML21eb, JWH25}.

In this work, we focus on linear regression models and the Tikhonov regularized estimator for ridge regression. It can also be interpreted as an empirical Bayes (EB) estimator, which is equipped with a Gaussian weighting function with a zero mean and an identity covariance matrix scaled by a scalar data-dependent hyper-parameter estimator. In this paper, we apply the EB hyper-parameter estimator \cite{PCCDL2022}. It corresponds to solving a non-convex optimization problem and has no closed-form solution.

The objective of this paper is to design novel hyper-parameter free estimators with performance close to that of the EB-based regularized estimator. The MSE is a common performance measure for estimators. However, the MSE of the EB-based regularized estimator is analytically intractable for finite sample sizes. To solve this issue, we have proposed an asymptotic criterion in \cite{JWH25}: the excess MSE (XMSE), which quantifies the difference in MSE between the EB-based regularized estimator and the ML estimator for large sample sizes. The interest in the XMSE stems from the fact that it can be computed analytically for many estimators \cite{JWH25}. In particular, we assume that the limit of the Gram matrix, whose entry is the inner product of columns of the regression matrix, is an identity matrix. For this setting, we will exploit the explicit XMSE expressions present in \cite[Theorem 2]{JWH25} to design new estimators that have the same XMSE as the EB-based regularized estimator. Our main contributions are:
\begin{enumerate}
\item We utilize the XMSE expression for generalized Bayes estimators given in \cite[Corollary 2.1]{JWH25} to develop a family of generalized Bayes estimators. They have the same XMSE as the EB-based regularized estimator, but being Bayes estimators, they do not depend on hyper-parameters that need to be estimated.
\item We derive an explicit XMSE expression for a biased estimator with a specific form. Based on this, we design a family of biased estimators in closed form, which have the same XMSE as the EB-based regularized estimator.
\item We conduct numerical simulations to show that the performance of these new estimators is comparable to that of the EB-based regularized estimator, while they are computationally more efficient.
\end{enumerate}

The remaining part of this paper is organized as follows. In Section \ref{sec: preliminaries}, we present preliminaries on the ML and regularized estimation for linear regression models and give the problem statement. In Section \ref{sec: Bayes estimator}, we provide a family of explicit weighting functions, and the corresponding generalized Bayes estimators have the same XMSE as the EB-based regularized estimator. In Section \ref{sec: Biased estimator}, we establish a family of biased estimators in closed form that have the same XMSE as the EB-based regularized estimator. In Section \ref{sec: numerical simulation}, we perform numerical experiments to show the average performance of the designed Bayes and biased estimators. In Section \ref{sec: conclusion}, we conclude this paper. The proofs of lemmas and theorems are included in Appendix A, and the required technical lemma is given in Appendix B. 

\textbf{Notation.} The set of real-valued matrices of dimension $m_{1}\times m_{2}$ is denoted ${\R}^{m_{1}\times m_{2}}$. The $m$-by-$m$ dimensional identity matrix is denoted $\mathbf{I}_{m}$. The first-order and second-order partial derivatives of $f(\bm{x}):\R^{m}\to \R$ with respect to $\bm{x}\in\R^{m}$ are denoted ${\partial f(\bm{x})}/{\partial \bm{x}}\in\R^{m}$ and ${\partial^2 f(\bm{x})}/{\partial \bm{x}\partial \bm{x}^{\top}}\in\R^{m\times m}$, respectively. The $(k,l)$th entry, the $k$th row and the $k$th column of a matrix $\mathbf{A}$ are $[\mathbf{A}]_{k,l}$, $[\mathbf{A}]_{k,:}$ and $[\mathbf{A}]_{:.k}$, respectively. The transpose and inverse of a matrix $\mathbf{A}$ are denoted $\mathbf{A}^{\top}$ and $\mathbf{A}^{-1}$, respectively. For a random vector $\bm{a}$, its expectation and covariance matrix are denoted $\E(\bm{a})$ and $\Var(\bm{a})\triangleq\E[(\bm{a}-\E(\bm{a}))(\bm{a}-\E(\bm{a}))^{\top}]$. The cross covariance matrix of random vectors $\bm{a}$ and $\bm{b}$ is denoted $\COV(\bm{a},\bm{b})\triangleq\E[(\bm{a}-\E(\bm{a}))(\bm{b}-\E(\bm{b}))^{\top}]$. A Gaussian distributed random variable $\bm\xi\in\R^{m}$ with mean $\bm{a}\in\R^{m}$ and covariance matrix $\mathbf{A}\in\R^{m\times m}$ is denoted $\mathcal{N}(\bm{a},\mathbf{A})$. The rank of a matrix is denoted $\rank(\cdot)$. The Euclidean norm of a vector is denoted $\|\cdot\|_{2}$. The trace and determinant of a square matrix are denoted $\Tr(\cdot)$ and $\det(\cdot)$, respectively. 

\section{Preliminaries and Problem Statement}\label{sec: preliminaries}


We consider the following linear regression model,
\begin{align}\label{eq: FIR model}
\bm{Y}=\mathbf{\Phi}\bm\theta+\bm{E},
\end{align}
where $\bm{Y}=[y(1),\cdots,y(N)]^{\top}$ is the measurement output vector with $N$ being the sample size, $\mathbf{\Phi}\in\R^{N\times n}$ is a lower triangular matrix consisting of inputs $\{u(t)\}_{t=1}^{N}$ and $[\mathbf{\Phi}]_{:,1}=[u(0),\cdots,u(N-1)]^{\top}$, and $\bm\theta\in\R^{n}$ is the model parameter vector to be estimated. The measurement noise vector $\bm{E}\in\R^{N}$ is assumed to be Gaussian distributed with zero mean and covariance $\sigma^2\mathbf{I}_{N}$, i.e., $\bm{E}\sim\mathcal{N}(\bm{0},\sigma^2\mathbf{I}_{N})$. Moreover, the inputs $\{u(t)\}_{t=1}^{N}$ are assumed to be known and deterministic with $u(t)=0$ for $t\leq 0$. Given an estimator $\hat{\bm\theta}\in\R^{n}$ of the unknown parameter $\bm\theta$, we evaluate its average performance by its mean squared error (MSE),
\begin{align}\label{eq: def of MSE}
\MSE(\hat{\bm\theta})=\E(\|\hat{\bm\theta}-\bm\theta_{0}\|_{2}^2),
\end{align}
where $\bm\theta_{0}\in\R^{n}$ is the ``true'' value of $\bm\theta$ and the expectation $\E$ is with respect to the measurement noise $\bm{E}$. The smaller its MSE, the better its performance.

\subsection{Maximum likelihood and regularized estimators}\label{subsec: ML and RLS}

Assume that the regression matrix $\mathbf{\Phi}$ in \eqref{eq: FIR model} has full column rank with $N\geq n$, i.e., $\rank(\mathbf{\Phi})=n$. One classical estimator of $\bm\theta$ is the ML estimator given by 
\begin{align}\label{eq: ML estimator}
\hat{\bm\theta}^{\ML}=&\argmin_{\bm\theta\in\R^{n}} p(\bm{Y}|\bm\theta)
=(\mathbf{\Phi}^{\top}\mathbf{\Phi})^{-1}\mathbf{\Phi}^{\top}\bm{Y},
\end{align}
where $p(\bm{Y}|\bm{\theta})$ is the probability density function (pdf) of $\bm{Y}|\bm\theta\sim\mathcal{N}(\mathbf{\Phi}\bm\theta,\sigma^2\mathbf{I}_{N})$. It is well-known that the ML estimator \eqref{eq: ML estimator} is unbiased but may have a large variance, which will result in large $\MSE(\hat{\bm\theta}^{\ML})=\sigma^2\Tr[(\mathbf{\Phi}^{\top}\mathbf{\Phi})^{-1}]$. 

To achieve a better bias-variance trade-off, we consider a weighting function $\pi(\bm\theta|\eta)$ as the pdf of $\bm\theta|\eta\sim\mathcal{N}(\bm{0},\eta\mathbf{I}_{n})$, where $\eta>0$ is known as the hyper-parameter and can be estimated from observations $\{u(t),y(t)\}_{t=1}^{N}$. Given a hyper-parameter estimator $\hat{\eta}$, the corresponding empirical Bayes (EB) model estimator is defined as
\begin{align}\label{eq: regularized estimator}
\hat{\bm\theta}^{\TR}(\hat{\eta})=&\argmin_{\hat{\bm\theta}\in\R^{n}}\int \|\hat{\bm\theta}-\bm\theta\|_{2}^2\frac{p(\bm{Y}|\bm{\theta})\pi(\bm\theta|\hat{\eta})}{\int p(\bm{Y}|\bm{\theta})\pi(\bm\theta|\hat{\eta})d\bm\theta}d\bm\theta\nonumber\\
=&[\mathbf{\Phi}^{\top}\mathbf{\Phi}+(\sigma^2/\hat{\eta})\mathbf{I}_{n}]^{-1}\mathbf{\Phi}^{\top}\bm{Y},
\end{align}
which is also called the regularized ridge regression estimator. One of the most commonly used hyper-parameter estimators is the EB hyper-parameter estimator \cite{PCCDL2022}:
\begin{subequations}\label{eq: EB hyper-parameter estimation}
\begin{align}\label{eq: EB hyper-parameter estimator}
\hat{\eta}_{\EB}=&\argmin_{\eta>0}\int p(\bm{Y}|\bm\theta)\pi(\bm\theta|\eta)d\bm\theta\nonumber\\
=&\argmin_{\eta>0}\mathscr{F}_{\EB}(\eta),\\
\label{eq: def of F_EB}
\mathscr{F}_{\EB}(\eta)=&\bm{Y}^{\top}\mathbf{Q}(\eta)^{-1}\bm{Y}+\log\det(\mathbf{Q}(\eta))
\end{align}
\end{subequations}
with $\mathbf{Q}(\eta)=\eta\mathbf{\Phi}\mathbf{\Phi}^{\top}+\sigma^2\mathbf{I}_{N}$. Correspondingly, $\hat{\bm\theta}^{\TR}(\hat{\eta}_{\EB})$ in the form of \eqref{eq: regularized estimator} will be referred to as the EB-based regularized estimator in this paper.

\subsection{Problem statement}\label{subsec: problem statement}

As discussed in \cite{MCL18asy, PDCDL2014}, the EB-based regularized estimator often performs better than the ML estimator. Although we can apply efficient algorithms, {see}, e.g., \cite{CL:12}, to compute the cost function $\mathscr{F}_{\EB}(\eta)$ in \eqref{eq: def of F_EB} and the corresponding regularized estimator \eqref{eq: regularized estimator}, the EB hyper-parameter estimation \eqref{eq: EB hyper-parameter estimation} corresponds to solving a non-convex optimization problem. The objective of this paper is to design new estimators that exhibit performance close to that of the EB-based regularized estimator but without hyper-parameter estimation.  

Since $\hat{\bm\theta}^{\TR}(\hat{\eta}_{\EB})$ is a nonlinear function of the measurement noise $\bm{E}$, it is hard to express $\MSE(\hat{\bm\theta}^{\TR}(\hat{\eta}_{\EB}))$ in closed form for finite sample sizes. As an alternative, we have proposed the excess MSE (XMSE) in \cite{JWH25}. It is an asymptotic criterion {that is analytically tractable and that} can be used to quantify the difference between $\MSE(\hat{\bm\theta}^{\TR}(\hat{\eta}_{\EB}))$ and $\MSE(\hat{\bm\theta}^{\ML})$ for large sample sizes. In \cite[Theorem 2]{JWH25}, we have derived an explicit expression for the XMSE of a general EB model estimator {under} the assumption that the limit of $\mathbf{\Phi}^{\top}\mathbf{\Phi}/N$ exists and is positive definite. In particular, the XMSE of $\hat{\bm\theta}^{\TR}(\hat{\eta}_{\EB})$ is defined as the limit of $N^2[\MSE(\hat{\bm\theta}^{\TR}(\hat{\eta}_{\EB}))-\MSE(\hat{\bm\theta}^{\ML})]$ as the sample size $N$ goes to infinity. As a special case of \cite[Theorem 2 and Corollaries 2.2-2.4]{JWH25}, the expression of $\XMSE(\hat{\bm\theta}^{\TR}(\hat{\eta}_{\EB}))$ is now revisited.

\begin{lemma}\label{lemma: XMSE expression of the EB-based regularized estimator}
   Assume that
   \begin{align}\label{eq: limit of regression matrix}
    \lim_{N\to\infty}\frac{\mathbf{\Phi}^{\top}\mathbf{\Phi}}{N}=\mathbf{I}_{n},
   \end{align}
   and moreover, $\hat{\eta}_{\EB}$ in \eqref{eq: EB hyper-parameter estimator} satisfies the first-order optimality condition, i.e., ${\partial \mathscr{F}_{\EB}(\eta)}/{\partial \eta}|_{\eta=\hat{\eta}_{\EB}}=0$, and its limit exists, i.e., $\lim_{N\to\infty}\hat{\eta}_{\EB}|_{\hat{\bm\theta}^{\ML}=\bm\theta_{0}}=\eta_{\tb\star}$. Then we have
   \begin{align}\label{eq: XMSE of thetaR eta_EB}
    \XMSE(\hat{\bm\theta}^{\TR}(\hat{\eta}_{\EB}))=\frac{(-n^2+4n)(\sigma^2)^2}{\|\bm\theta_{0}\|_{2}^2}.
   \end{align}
\end{lemma}

For large sample sizes, Lemma \ref{lemma: XMSE expression of the EB-based regularized estimator} implies
\begin{itemize}
\item[-] for $n=1,\cdots,4$, $\XMSE(\hat{\bm\theta}^{\TR}(\hat{\eta}_{\EB}))\geq 0$, which means that the performance of $\hat{\bm\theta}^{\ML}$ is no worse than that of $\hat{\bm\theta}^{\TR}(\hat{\eta}_{\EB})$ and there is no need {for} regularization;
\item[-] for $n=5,\cdots$, $\XMSE(\hat{\bm\theta}^{\TR}(\hat{\eta}_{\EB}))<0$, which means that $\hat{\bm\theta}^{\TR}(\hat{\eta}_{\EB})$ will outperform $\hat{\bm\theta}^{\ML}$. Notice that for fixed $\sigma^2$ and $\|\bm\theta_{0}\|_{2}$, the larger $n$, the more {significant} the improvement of $\hat{\bm\theta}^{\TR}(\hat{\eta}_{\EB})$ over $\hat{\bm\theta}^{\ML}$ becomes.
\end{itemize}

The closed-form expression for $\XMSE(\hat{\bm\theta}^{\TR}(\hat{\eta}_{\EB}))$ in \eqref{eq: XMSE of thetaR eta_EB} enables us to explore the possibilities of new estimators that have the same XMSE as {$\hat{\bm\theta}^{\TR}(\hat{\eta}_{\EB})$}. To be more specific, we focus on the following two problems.
\begin{enumerate}
\item Is it possible to design a generalized Bayes estimator\footnote{For a generalized Bayes estimator, its nonnegative weighting function $\pi(\bm\theta)$ can be improper, i.e., $\int \pi(\bm\theta)d\bm\theta=+\infty$.} defined as
\begin{align}\label{eq: def of Bayes estimator}
\hat{\bm\theta}^{\Bayes}=&\argmin_{\hat{\bm\theta}\in\R^{n}}\int \|\hat{\bm\theta}-\bm\theta\|_{2}^2\frac{p(\bm{Y}|\bm{\theta})\pi(\bm\theta)}{\int p(\bm{Y}|\bm{\theta})\pi(\bm\theta)d\bm\theta}d\bm\theta\nonumber\\
=&\frac{\int \bm\theta p(\bm{Y}|\bm{\theta})\pi(\bm\theta)d\bm\theta}{\int p(\bm{Y}|\bm{\theta})\pi(\bm\theta)d\bm\theta},
\end{align}
that has the same XMSE as $\hat{\bm\theta}^{\TR}(\hat{\eta}_{\EB})$? Clearly, $\hat{\bm\theta}^{\Bayes}$ in \eqref{eq: def of Bayes estimator} {is free of hyper-parameters, thereby eliminating the computational cost associated with estimating such parameters}.
\item Although a generalized Bayes estimator in \eqref{eq: def of Bayes estimator} does not need any hyper-parameter, it often has no closed-form expression and thus needs to be computed using sampling methods. The question thus arises whether it is possible to design a biased estimator in closed form that has the same XMSE as $\hat{\bm\theta}^{\TR}(\hat{\eta}_{\EB})$.
\end{enumerate} 

In the following, we will solve these two problems and evaluate the performance of the new estimators numerically.

\section{Bayes Estimators with the Same XMSE as the EB-based Regularized Estimator}\label{sec: Bayes estimator}

Note that the XMSE expression of a generalized Bayes estimator $\hat{\bm\theta}^{\Bayes}$ in \eqref{eq: def of Bayes estimator} has been given in \cite[Corollary 2.1]{JWH25}. 

\begin{lemma}\label{lemma: XMSE of a generalized Bayes estimator}
Assume that \eqref{eq: limit of regression matrix} holds. Then we have
\begin{align}\label{eq: XMSE of a generalized Bayes estimator}
\XMSE(\hat{\bm\theta}^{\Bayes})=&\left.-\frac{(\sigma^2)^2}{[\pi(\bm\theta)]^2}\left(\frac{\partial \pi(\bm\theta)}{\partial \bm{\theta}}\right)^{\top}\frac{\partial \pi(\bm\theta)}{\partial \bm{\theta}}\right|_{\bm\theta=\bm\theta_{0}}\nonumber\\
&+\left.\frac{2(\sigma^2)^2}{\pi(\bm\theta)}\Tr\left[\frac{\partial^2 \pi(\bm\theta)}{\partial \bm\theta\partial\bm\theta^{\top}} \right]\right|_{\bm\theta=\bm\theta_{0}}.
\end{align}
\end{lemma}

Based on Lemmas \ref{lemma: XMSE expression of the EB-based regularized estimator}-\ref{lemma: XMSE of a generalized Bayes estimator}, we can design a family of explicit weighting functions $\pi(\bm\theta)$ such that the corresponding generalized Bayes estimators have the same XMSE as $\hat{\bm\theta}^{\TR}(\hat{\eta}_{\EB})$.

\begin{theorem}\label{thm: explicit weighting function}
Under the assumptions of Lemma \ref{lemma: XMSE expression of the EB-based regularized estimator}, if the weighting function of $\hat{\bm\theta}^{\Bayes,\EB}$ is in the form of 
\begin{align}\label{eq: prior family EB}
\pi(\bm\theta)=\|\bm\theta\|_{2}^{2-n}(C_{1}\|\bm\theta\|_{2}+C_{2}\|\bm\theta\|_{2}^{-1})^2,
\end{align}
where $C_{1},C_{2}\in\R$ are arbitrary constants, we then have
\begin{align}\label{eq: identical XMSE of Bayes_EB estimator}
\XMSE(\hat{\bm\theta}^{\Bayes,\EB})=\XMSE(\hat{\bm\theta}^{\TR}(\hat{\eta}_{\EB})).
\end{align}
\end{theorem}

\begin{remark}
To ensure the weighting function \eqref{eq: prior family EB} to be well-defined, we can replace $\|\bm\theta\|_{2}$ by $\|\bm\theta+\delta\|_{2}$ in applications, where $\delta$ is a small positive number.
\end{remark}

Theorem \ref{thm: explicit weighting function} provides a family of generalized Bayes estimators that have the same XMSE as the EB-based regularized estimator. They have no closed-form expressions and need to be implemented using sampling methods. Here, we utilize the direct sampling approximation of \cite[(A.2)]{JWH25}, which corresponds to generating $\{\bm\theta_{k}\}_{k=1}^{M_s}$ as $M_{s}$ independent realizations of $\mathcal{N}(\hat{\bm\theta}^{\ML},\sigma^2(\mathbf{\Phi}^{\top}\mathbf{\Phi})^{-1})$ and calculate 
   \begin{align}\label{eq: sample approximation of Bayes estimator}
    \hat{\bm\theta}^{\Bayes}\approx
   \sum_{k=1}^{M_s}\frac{\bm\theta_{k}\pi(\bm\theta_{k})}{\sum_{k=1}^{M_s}{\pi(\bm\theta_{k})}}.
   \end{align}
With $M_{s}$ increasing, the approximation \eqref{eq: sample approximation of Bayes estimator} will be more accurate, while its computation will be more expensive. To avoid this, we will now consider how to design closed-form biased estimators that have the same XMSE as $\hat{\bm\theta}^{\TR}(\hat{\eta}_{\EB})$ in the next section.

\section{Biased Estimators with the Same XMSE as the EB-based Regularized Estimator}\label{sec: Biased estimator}

We consider a biased estimator in the form of
\begin{align}\label{eq: def of biased estimator}
\hat{\bm\theta}^{\Biased}=\hat{\bm\theta}^{\ML}+\frac{1}{N}\bm{b}_{N}(\hat{\bm\theta}^{\ML}),
\end{align}
where $\bm{b}_{N}(\hat{\bm\theta}^{\ML})\in\R^{n}$ is a function of $\hat{\bm\theta}^{\ML}$. The XMSE of $\hat{\bm\theta}^{\Biased}$ is given in the following theorem.

\begin{theorem}\label{thm: XMSE of biased estimator}
Assume that \eqref{eq: limit of regression matrix} holds, and that the following two limits exist,
\begin{subequations}\label{eq: limits of b_N and its derivative}\noeqref{eq: limit of b_N}\noeqref{eq: limit of 1st order derivative of b_N}
 \begin{gather}\label{eq: limit of b_N}
 \lim_{N\to\infty}\left.\bm{b}_{N}(\hat{\bm\theta}^{\ML})\right|_{\hat{\bm\theta}^{\ML}=\bm\theta_{0}}=\bm{b}_{\star}(\bm\theta_{0}),\\
 \label{eq: limit of 1st order derivative of b_N}
 \lim_{N\to\infty}\left.\frac{\partial \bm{b}_{N}(\hat{\bm\theta}^{\ML})}{\partial \hat{\bm\theta}^{\ML}}\right|_{\hat{\bm\theta}^{\ML}=\bm\theta_{0}}=\mathbf{b}_{\star}^{'}(\bm\theta_{0}).
 \end{gather}
\end{subequations}
Then, the XMSE of $\hat{\bm\theta}^{\Biased}$ can be expressed as
\begin{align}\label{eq: XMSE of biased estimator}
&\XMSE(\hat{\bm\theta}^{\Biased})=\nonumber\\
&\|\XBias(\hat{\bm\theta}^{\Biased})\|_{2}^2+\Tr[\XVar(\hat{\bm\theta}^{\Biased})],
\end{align}
where
\begin{align*}
\XBias(\hat{\bm\theta}^{\Biased})=&\bm{b}_{\star}(\bm\theta_{0}),\\
\XVar(\hat{\bm\theta}^{\Biased})=&{\XVar}_{\#}(\hat{\bm\theta}^{\Biased})+{\XVar}_{\#}(\hat{\bm\theta}^{\Biased})^{\top},\\
{\XVar}_{\#}(\hat{\bm\theta}^{\Biased})=&\sigma^2\mathbf{b}_{\star}^{'}(\bm\theta_{0}).
\end{align*}
\end{theorem}

As shown in \eqref{eq: XMSE of biased estimator}, $\XMSE(\hat{\bm\theta}^{\Biased})$ consists of the two components: $\|\XBias(\hat{\bm\theta}^{\Biased})\|_{2}^2$ and $\Tr[\XVar(\hat{\bm\theta}^{\Biased})]$, which are denoted the excess squared bias and the excess variance of $\hat{\bm\theta}^{\Biased}$, respectively. Based on Theorems \ref{thm: explicit weighting function}-\ref{thm: XMSE of biased estimator}, we can design explicit forms of $\bm{b}_{N}(\hat{\bm\theta}^{\ML})$ and the corresponding biased estimators will have the same XMSE as $\hat{\bm\theta}^{\TR}(\hat{\eta}_{\EB})$ and $\hat{\bm\theta}^{\Bayes,\EB}$.



\begin{theorem}\label{thm: explicit form of biased term}
Let the assumptions of Lemma \ref{lemma: XMSE expression of the EB-based regularized estimator} hold. If $\hat{\bm\theta}^{\Biased,\EB}$ takes the form of \eqref{eq: def of biased estimator} with
\begin{align}\label{eq: explicit form of bias term}
&\bm{b}_{N}(\hat{\bm\theta}^{\ML})=\sigma^2N(\mathbf{\Phi}^{\top}\mathbf{\Phi})^{-1}\nonumber\\
&\times \left[ 2-n+\frac{2(C_{1}\|\hat{\bm\theta}^{\ML}\|_{2}-C_{2}\|\hat{\bm\theta}^{\ML}\|_{2}^{-1})}{C_{1}\|\hat{\bm\theta}^{\ML}\|_{2}+C_{2}\|\hat{\bm\theta}^{\ML}\|_{2}^{-1}}\right]\frac{\hat{\bm\theta}^{\ML}}{\|\hat{\bm\theta}^{\ML}\|_{2}^2},\quad\  
\end{align}
where $C_{1},C_{2}\in\R$ are arbitrary constants, we then have
\begin{align*}
\XMSE(\hat{\bm\theta}^{\Biased,\EB})=&\XMSE(\hat{\bm\theta}^{\Bayes,\EB})\\
=&\XMSE(\hat{\bm\theta}^{\TR}(\hat{\eta}_{\EB})).
\end{align*}
\end{theorem}

\begin{remark}
Note that $\hat{\bm\theta}^{\Bayes,\EB}$ with \eqref{eq: prior family EB} and $\hat{\bm\theta}^{\Biased,\EB}$ with \eqref{eq: explicit form of bias term} are both families of estimators indexed by $C_{1}$ and $C_{2}$. From the expression of a generalized Bayes estimator in \eqref{eq: def of Bayes estimator}, and the explicit forms of $\pi(\bm\theta)$ in \eqref{eq: prior family EB} and $\bm{b}_{N}(\hat{\bm\theta}^{\ML})$ in \eqref{eq: explicit form of bias term}, the performance of $\hat{\bm\theta}^{\Bayes,\EB}$ and $\hat{\bm\theta}^{\Biased,\EB}$ depends only on the ratio of $C_{1}$ and $C_{2}$, rather than their individual values.
\end{remark}

In Theorem \ref{thm: explicit form of biased term}, we present a family of closed-form biased estimators with the same XMSE as $\hat{\bm\theta}^{\TR}(\hat{\eta}_{\EB})$. Moreover, we can slightly generalize the results above.

\begin{theorem}\label{thm: generalization of conditions}
If the assumptions of Lemma \ref{lemma: XMSE expression of the EB-based regularized estimator} hold but with \eqref{eq: limit of regression matrix} replaced by $\lim_{N\to\infty}{\mathbf{\Phi}^{\top}\mathbf{\Phi}}/{N}=\sigma_{u}^2\mathbf{I}_{n}$, then Theorems \ref{thm: explicit weighting function} and \ref{thm: explicit form of biased term} remain true.
\end{theorem}

\section{Numerical Simulation}\label{sec: numerical simulation}

In this section, we conduct numerical simulations to evaluate the performance of $\hat{\bm\theta}^{\Bayes,\EB}$ with \eqref{eq: prior family EB} and $\hat{\bm\theta}^{\Biased,\EB}$ in the form of \eqref{eq: def of biased estimator} with \eqref{eq: explicit form of bias term}.

\subsection{Test systems and data}

We will generate $100$ collections of test systems and input-output data. For each collection, we
\begin{enumerate}
\item generate $\tilde{\bm\theta}_{0}$ as a realization of $\mathcal{N}(\bm{0},\mathbf{I}_{n})$ and scale $\bm\theta_{0}=m_{\theta}\tilde{\bm\theta}_{0}$ such that $\|\bm\theta_{0}\|_{2}=1$;
\item generate $\{\tilde{u}(t)\}_{t=1}^{N}$ as independent realizations of $\mathcal{N}(0,1)$ and set $\sigma^2=1$;
\item scale $u(t)=m_{u}\tilde{u}(t)$ such that the sample SNR, which is the ratio between the sample variance of $\mathbf{\Phi}\bm\theta_{0}$ and the measurement noise variance $\sigma^2$, is $5$;
\item corrupt the noise-free output $\mathbf{\Phi}\bm\theta_{0}$ with $N_{\text{MC}}=200$ additive independent noise realizations of $\mathcal{N}(\bm{0},\sigma^2\mathbf{I}_{n})$, to obtain $N_{\text{MC}}$ measurement output sequences $\{y(t)\}_{t=1}^{N}$.
\end{enumerate}

\subsection{Simulation setup}

For each collection of test system and input-output data, we will perform $N_{\text{MC}}=200$ Monte Carlo (MC) simulations to demonstrate the average performance of $\hat{\bm\theta}^{\ML}$, $\hat{\bm\theta}^{\TR}(\hat{\eta}_{\EB})$, $\hat{\bm\theta}^{\Bayes,\EB}$ with \eqref{eq: prior family EB} and $\hat{\bm\theta}^{\Biased,\EB}$ with \eqref{eq: explicit form of bias term} for different selections of $n, N, C_{1}, C_{2}$. The regularized estimator $\hat{\bm\theta}^{\TR}(\hat{\eta}_{\EB})$ will be implemented using \cite[Algorithm 2]{CL:12}. The Bayes estimator $\hat{\bm\theta}^{\Bayes,\EB}$ will be approximated using \eqref{eq: sample approximation of Bayes estimator}. The average performance of an estimator $\hat{\bm\theta}$ will be measured by the sample mean of $\|\hat{\bm\theta}-\bm\theta_{0}\|_{2}$ over $200$ MC simulations, referred to as the sample $\MSE(\hat{\bm\theta})$. As its relative version, the average $\FIT(\hat{\bm\theta})$ \cite{Ljung1995} is given by the sample mean of
\begin{align}
\FIT(\hat{\bm\theta})=100\times \left(1-\frac{\|\hat{\bm\theta}-\bm\theta_{0}\|_{2}}{\|\bm\theta_{0}-\bar{\theta}_{0}\|_{2}} \right),\ \bar{\theta}_{0}=\frac{1}{n}\sum_{k=1}^{n}[\bm\theta_{0}]_{k},
\end{align}
over $200$ MC simulations. The better $\hat{\bm\theta}$ performs, the smaller its sample MSE, while the larger its average FIT. To better display the average performance of $\hat{\bm\theta}^{\ML}$, $\hat{\bm\theta}^{\TR}(\hat{\eta}_{\EB})$, $\hat{\bm\theta}^{\Bayes,\EB}$ and $\hat{\bm\theta}^{\Biased,\EB}$, we will present the sample means of both the sample MSE and the average FIT of these estimators over $100$ collections of test systems and input-output data.

\subsection{Simulation results}

We consider the following two settings:
\begin{enumerate}
\item $n=1$ $N=5$, and $M_{s}=200$;
\item $n=5$, $N=15$, and $M_{s}=500$.
\end{enumerate}
From Fig. \ref{fig: identity Sigma and ridge regression for n=1, N=5 (Bayes and Biased)}-\ref{fig: identity Sigma and ridge regression for n=5, N=15 (Bayes and Biased)}, we can observe that
\begin{itemize}
\item[-] for $n=1, N=5$, $\hat{\bm\theta}^{\ML}$ outperforms $\hat{\bm\theta}^{\TR}(\hat{\eta}_{\EB})$, while for $n=5, N=15$, the performance of $\hat{\bm\theta}^{\ML}$ is worse, which confirms the discussions after Lemma \ref{lemma: XMSE expression of the EB-based regularized estimator};
\item[-] for at least one combination of $C_{1}$ and $C_{2}$, $\hat{\bm\theta}^{\Bayes,\EB}$ and $\hat{\bm\theta}^{\Biased,\EB}$ perform similarly to $\hat{\bm\theta}^{\TR}(\hat{\eta}_{\EB})$;
\item[-] among different combinations of $C_{1}$ and $C_{2}$, for $n=1$, $\hat{\bm\theta}^{\Bayes,\EB}$ and $\hat{\bm\theta}^{\Biased,\EB}$ with $C_{1}=1, C_{2}=0$ perform the best; while for $n=5$, $\hat{\bm\theta}^{\Bayes,\EB}$ and $\hat{\bm\theta}^{\Biased,\EB}$ with $C_{1}=0, C_{2}=1$ perform the best. 
\end{itemize}

\begin{figure}[!hbtp]
\centering
\includegraphics[width=\linewidth]{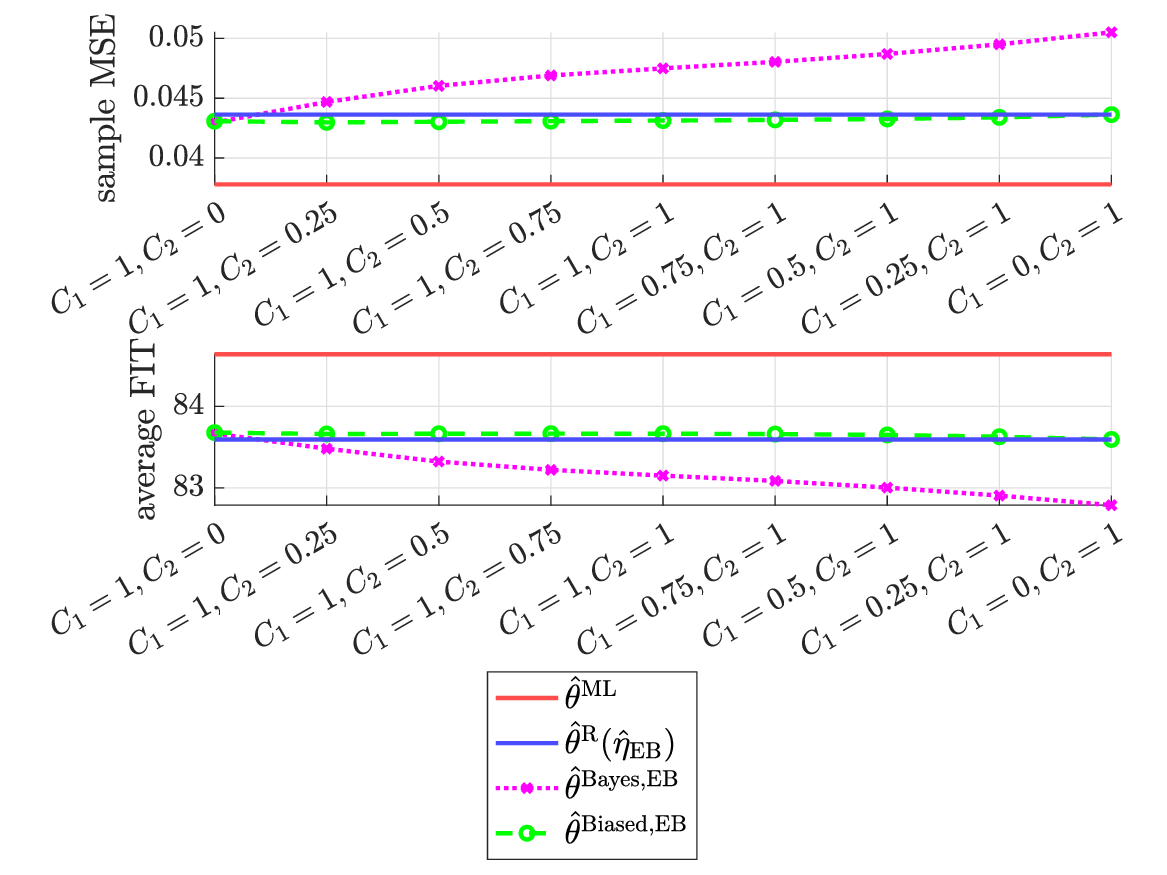}
\caption{Sample means of the sample MSE and the average FIT of $\hat{\bm\theta}^{\ML}$, $\hat{\bm\theta}^{\TR}(\hat{\eta}_{\EB})$, $\hat{\bm\theta}^{\Bayes,\EB}$ and $\hat{\bm\theta}^{\Biased,\EB}$ for $n=1$ and $N=5$.}
\label{fig: identity Sigma and ridge regression for n=1, N=5 (Bayes and Biased)}
\end{figure}

\begin{figure}[!hbtp]
\centering
\includegraphics[width=\linewidth]{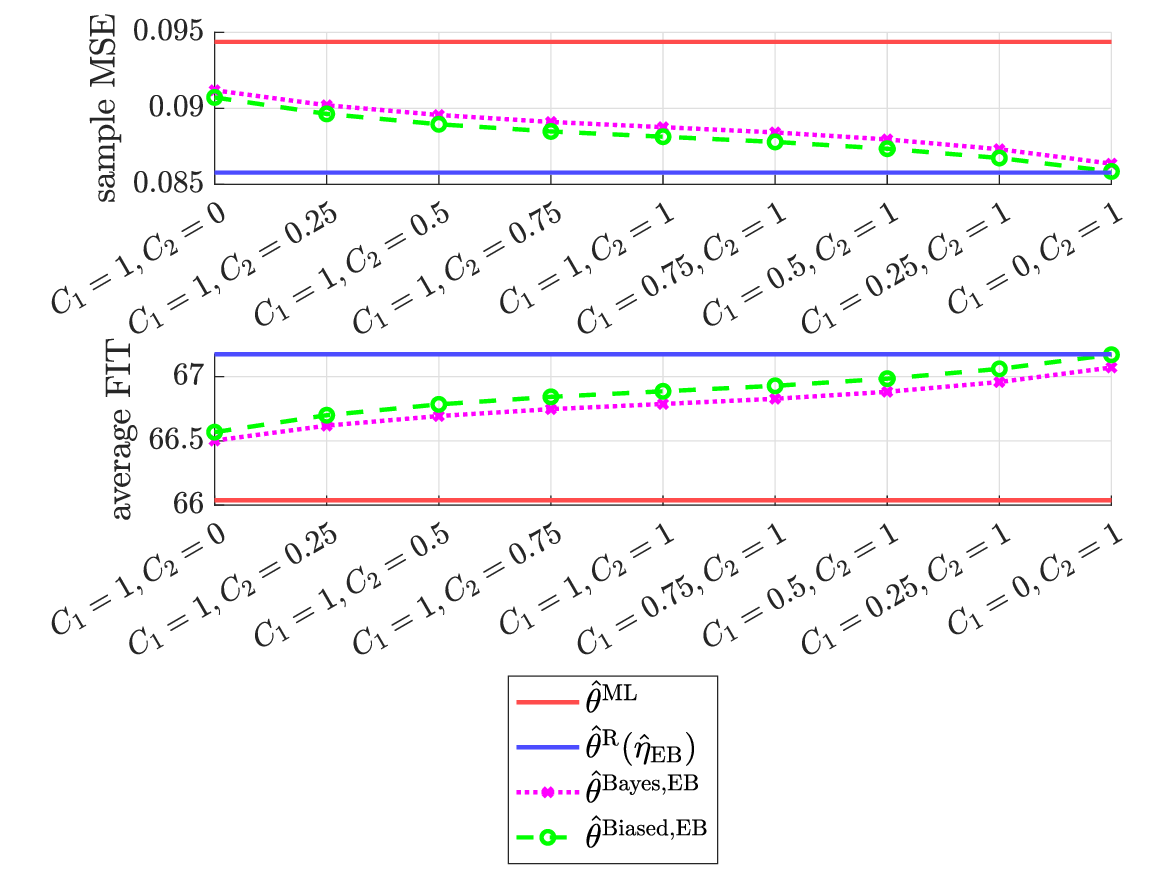}
\caption{Sample means of the sample MSE and the average FIT of $\hat{\bm\theta}^{\ML}$, $\hat{\bm\theta}^{\TR}(\hat{\eta}_{\EB})$, $\hat{\bm\theta}^{\Bayes,\EB}$ and $\hat{\bm\theta}^{\Biased,\EB}$ for $n=5$ and $N=15$.}
\label{fig: identity Sigma and ridge regression for n=5, N=15 (Bayes and Biased)}
\end{figure}

For larger $n$ and $N$, the influence of different ratios of $C_{1}$ and $C_{2}$ on the performance of $\hat{\bm\theta}^{\Bayes,\EB}$ and $\hat{\bm\theta}^{\Biased,\EB}$ becomes weaker. In Fig. \ref{fig: identity Sigma and ridge regression for n=80, N=360 (Bayes and Biased)} and Table\footnote{For $\hat{\bm\theta}^{\Bayes,\EB}$ and $\hat{\bm\theta}^{\Biased,\EB}$, we first calculate the total computing time, and the sample means of the average FIT and the sample MSE over $100$ collections of test systems and data. Then, we calculate the sample means of these three statistics over different combinations of $C_{1}$ and $C_{2}$.}~\ref{table: Bayes and Biased for n=80, N=360}, we consider $n=80$, $N=360$, and $M_{s}=5\times 10^{3}$. We can observe that the performance of $\hat{\bm\theta}^{\TR}(\hat{\eta}_{\EB})$ is quite close to that of $\hat{\bm\theta}^{\Bayes,\EB}$ and $\hat{\bm\theta}^{\Biased,\EB}$, while its computing time is over twice that of $\hat{\bm\theta}^{\Bayes,\EB}$ and over $500$ times that of $\hat{\bm\theta}^{\Biased,\EB}$.

\begin{figure}[!hbtp]
\centering
\includegraphics[width=\linewidth]{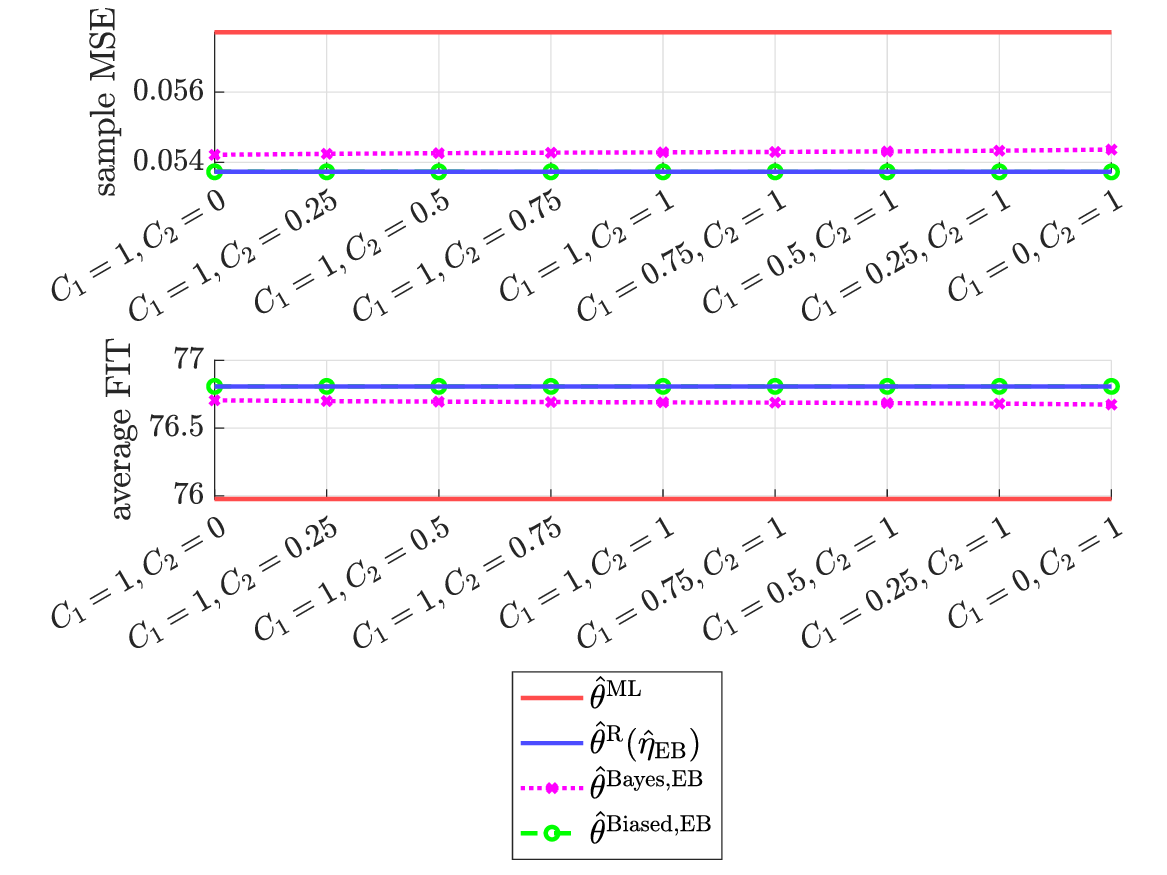}
\caption{Sample means of the sample MSE and the average FIT of $\hat{\bm\theta}^{\ML}$, $\hat{\bm\theta}^{\TR}(\hat{\eta}_{\EB})$, $\hat{\bm\theta}^{\Bayes,\EB}$ and $\hat{\bm\theta}^{\Biased,\EB}$ for $n=80$ and $N=360$.}
\label{fig: identity Sigma and ridge regression for n=80, N=360 (Bayes and Biased)}
\end{figure}

\begin{table}[!htbp]
\centering
\caption{Sample means of the sample MSE, the average FIT, and the total computing time of $\hat{\bm\theta}^{\TR}(\hat{\eta}_{\EB})$, $\hat{\bm\theta}^{\Bayes,\EB}$ and $\hat{\bm\theta}^{\Biased,\EB}$ for $n=80$ and $N=360$.}
\label{table: Bayes and Biased for n=80, N=360}
\resizebox{0.98\hsize}{!}{
\begin{tabular}{cccc}
\hline
& $\hat{\bm\theta}^{\TR}(\hat{\eta}_{\EB})$ & $\hat{\bm\theta}^{\Bayes,\EB}$ & $\hat{\bm\theta}^{\Biased,\EB}$ \\
\hline
sample MSE & $5.37\times 10^{-2}$ & $5.43\times 10^{-2}$ & $5.37\times 10^{-2}$\\
average FIT & $76.81$ & $76.69$ & $76.81$ \\
computing time (s) & $1.81\times 10^{3}$ & $6.68\times 10^{2}$ & $3.51$\\
\hline
\end{tabular}}
\end{table}

\section{Conclusion}\label{sec: conclusion}

This paper has introduced novel estimators that achieve performance comparable to the empirical Bayes (EB)-based regularized estimator for ridge regression but without hyper-parameter estimation. We assumed that, as the sample size goes to infinity, the limit of the Gram matrix is an identity matrix. Under this assumption, we utilized the excess MSE (XMSE) proposed in \cite{JWH25} to quantify the performance of estimators for large sample sizes. By exploiting the XMSE expressions in \cite{JWH25}, we have designed a family of generalized Bayes estimators that have the same XMSE as the EB-based regularized estimator. Moreover, we have derived an explicit XMSE expression for a class of biased estimators. Based on this, we have developed a family of closed-form biased estimators with the same XMSE as the EB-based regularized estimator. Numerical results have demonstrated that these estimators achieve performance comparable to the EB-based regularized estimator but are more computationally efficient. In future work, we will extend the present contributions into methods for designing hyper-parameter free estimators that match or exceed the performance of EB-based regularized estimators for a variety of kernels while being computationally advantageous.




\def\thesectiondis{\thesection.}                   
\def\thesubsectiondis{\thesection.\arabic{subsection}.}          
\def\thesubsubsectiondis{\thesubsection.\arabic{subsubsection}.}

\setcounter{subsection}{0}

\renewcommand{\thesection}{A}
\setcounter{lemma}{0}
\renewcommand{\thelemma}{A.\arabic{lemma}}

\renewcommand{\theequation}{A.\arabic{equation}}
\setcounter{equation}{0}

\renewcommand{\thesubsection}{\thesection.\arabic{subsection}}

\section*{Appendix A}\label{sec:Appendix A}

Proofs of lemmas and theorems are given in this section.

\subsection{Proof of Lemma \ref{lemma: XMSE expression of the EB-based regularized estimator}}

From \cite[Corollary 2.4]{JWH25}, we have $\eta_{\tb\star}={\bm\theta_{0}^{\top}\bm\theta_{0}}/{n}$.
Then we use \cite[Theorem 2 and Corollaries 2.2-2.4]{JWH25} to derive
\begin{gather*}
\|\XBias(\hat{\bm\theta}^{\TR}(\hat{\eta}_{\EB}))\|_{2}^2={n^2(\sigma^2)^2}/{\|\bm\theta_{0}\|_{2}^2},\\
\Tr[\XVar(\hat{\bm\theta}^{\TR}(\hat{\eta}_{\EB}))]=-{2n^2(\sigma^2)^2}/{\|\bm\theta_{0}\|_{2}^2},\\
\Tr[\XVarHPE(\hat{\bm\theta}^{\TR}(\hat{\eta}_{\EB}))]={4n(\sigma^2)^2}/{\|\bm\theta_{0}\|_{2}^2}.
\end{gather*}
Note that $\XMSE(\hat{\bm\theta}^{\TR}(\hat{\eta}_{\EB}))$ is the sum of these three terms.

\subsection{Proof of Lemma \ref{lemma: XMSE of a generalized Bayes estimator}}\label{subsec: Proof of XMSE of Bayes estimator}

We recall from \cite[Corollary 2.1]{JWH25} that 
\begin{align*}
&\|\XBias(\hat{\bm\theta}^{\Bayes})\|_{2}^2=(\sigma^2)^2\left\|\left. \frac{\partial \log(\pi(\bm\theta))}{\partial \bm\theta}\right|_{\bm\theta=\bm\theta_{0}}\right\|_{2}^2,\\
&\Tr[\XVar(\hat{\bm\theta}^{\Bayes})]=2(\sigma^2)^2\Tr\left[\left. \frac{\partial^2 \log(\pi(\bm\theta))}{\partial \bm\theta \partial \bm\theta^{\top}}\right|_{\bm\theta=\bm\theta_{0}}\right],\quad 
\end{align*}
and $\XMSE(\hat{\bm\theta}^{\Bayes})$ is the sum of these two terms, where
\begin{align}\label{eq:D1 of log prior}
\frac{\partial \log(\pi(\bm\theta))}{\partial \bm\theta}=&\frac{1}{\pi(\bm\theta)}\frac{\partial \pi(\bm\theta)}{\partial \bm\theta},\\
\frac{\partial^2 \log(\pi(\bm\theta))}{\partial \bm\theta\partial \bm\theta^{\top}}
\label{eq:D2 of log prior}
=&-\frac{1}{\pi(\bm\theta)^2}\frac{\partial \pi(\bm\theta)}{\partial \bm\theta}\frac{\partial \pi(\bm\theta)}{\partial \bm\theta^{\top}}+\frac{1}{\pi(\bm\theta)}\frac{\partial^2 \pi(\bm\theta)}{\partial \bm\theta\partial \bm\theta^{\top}}.
\end{align}
Hence, we obtain \eqref{eq: XMSE of a generalized Bayes estimator}.

\subsection{Proof of Theorem \ref{thm: explicit weighting function}}

We first let \eqref{eq: XMSE of thetaR eta_EB} equal \eqref{eq: XMSE of a generalized Bayes estimator} to obtain the following condition
 \begin{align*}
-\frac{1}{\pi(\bm\theta)^2}\left\|\frac{\partial \pi(\bm\theta)}{\partial \bm{\theta}}\right\|_{2}^2
+\frac{2}{\pi(\bm\theta)}\Tr\left[\frac{\partial^2 \pi(\bm\theta)}{\partial \bm\theta\partial\bm\theta^{\top}} \right]
=\frac{-n^2+4n}{\|\bm\theta\|_{2}^2}.
\end{align*}
Then we consider $\pi(\bm\theta)$ as a function of $\|\bm\theta\|_{2}$ and define $\pi(\bm\theta)=f(r)> 0$ with $r=\|\bm\theta\|_{2}$. Let $f_{r}^{'}$ and $f^{''}_{r}$ denote the first-order and second-order derivatives of $f(r)$ with respect to $r$, respectively. Note that
\begin{align*}
\frac{\partial \pi(\bm\theta)}{\partial \bm\theta}
=f^{'}_{r}\frac{\bm\theta}{\|\bm\theta\|_{2}},\
\Tr\left[\frac{\partial^2 \pi(\bm\theta)}{\partial \bm\theta\partial\bm\theta^{\top}}\right]=f_{r}^{''}+f_{r}^{'}\frac{n-1}{r}.
\end{align*}
Then, we can rewrite the condition above as
\begin{align*}
f(r)f_{r}^{''}-\frac{1}{2}(f_{r}^{'})^2+\frac{n-1}{r}f(r)f_{r}^{'}+\frac{n^2-4n}{2r^2}f(r)^2=0.
\end{align*}
Define $w(r)=f(r)^{1/2}$. It follows that $f(r)=w(r)^2$, $f_{r}^{'}=2w(r)w_{r}^{'}$ and $f_{r}^{''}=2(w_{r}^{'})^2+2w(r)w_{r}^{''}$. Then, we have
\begin{align*}
r^2w_{r}^{''}+(n-1)rw_{r}^{'}+[{(n^2-4n)}/{4}]w(r)=0,
\end{align*}
which is a second-order Euler equation in the form of \eqref{eq: second-order Euler equation}. By using Lemma \ref{lemma: second-order Euler equation}, we have solutions of $w(r)$ and $\pi(\bm\theta)$.

\subsection{Proof of Theorem \ref{thm: XMSE of biased estimator}}

We start by decomposing $\MSE(\hat{\bm\theta}^{\Biased})-\MSE(\hat{\bm\theta}^{\ML})$ as the sum of $\|\bm{\Upsilon}_{\text{Bias}}\|_{2}^2$, $\Tr(\mathbf{\Upsilon}_{\text{Var}})$ and $\Tr(\mathbf{\Upsilon}_{\text{HOT}})$,
where $\bm{\Upsilon}_{\text{Bias}}=\E(\hat{\bm\theta}^{\Biased})-\bm\theta_{0}=\E(\hat{\bm\theta}^{\Biased}-\hat{\bm\theta}^{\ML})$, $\mathbf{\Upsilon}_{\text{Var}}=\COV(\hat{\bm\theta}^{\Biased}-\hat{\bm\theta}^{\ML},\hat{\bm\theta}^{\ML})+\COV(\hat{\bm\theta}^{\ML},\hat{\bm\theta}^{\Biased}-\hat{\bm\theta}^{\ML})$,
and $\mathbf{\Upsilon}_{\text{HOT}}=\Var(\hat{\bm\theta}^{\Biased}-\hat{\bm\theta}^{\ML})$. Then we derive the first-order Taylor expansion\footnote{Note that for random sequences $\xi_{N}\in\C$ and $a_{N}\in\C$, if $\xi_{N}/a_{N}$ converges in probability to $0$ as $N\to\infty$, we denote it as $\xi_{N}=o_{p}(a_{N})$.} to $\hat{\bm\theta}^{\Biased}-\hat{\bm\theta}^{\ML}$ at $\hat{\bm\theta}^{\ML}=\bm\theta_{0}$,
\begin{align*}
&\hat{\bm\theta}^{\Biased}-\hat{\bm\theta}^{\ML}
=({1}/{N})\bm{b}_{N}(\bm\theta_{0})\\
&+({1}/{N})[{\partial \bm{b}_{N}(\hat{\bm\theta}^{\ML})}/{\partial \hat{\bm\theta}^{\ML}}|_{\hat{\bm\theta}^{\ML}=\bm\theta_{0}}](\hat{\bm\theta}^{\ML}-\bm\theta_{0})\\
&+o_{p}(\|\hat{\bm\theta}^{\ML}-\bm\theta_{0}\|_{2}),
\end{align*}
where $\bm{b}_{N}(\bm\theta_{0})=\bm{b}_{N}(\hat{\bm\theta}^{\ML})|_{\hat{\bm\theta}^{\ML}=\bm\theta_{0}}$. It follows that\footnote{Note that for sequences $\xi_{N}\in\C$ and $a_{N}\in\C$, if $\lim_{N\to\infty}\xi_{N}/a_{N}=0$, we denote it as $\xi_{N}=o(a_{N})$.} 
\begin{align*}
&\E(\hat{\bm\theta}^{\Biased}-\hat{\bm\theta}^{\ML})=({1}/{N})\bm{b}_{N}(\bm\theta_{0})+o(1/N),\\
&\COV(\hat{\bm\theta}^{\Biased}-\hat{\bm\theta}^{\ML},\hat{\bm\theta}^{\ML})=\\
&({1}/{N})\sigma^2[{\partial \bm{b}_{N}(\hat{\bm\theta}^{\ML})}/{\partial \hat{\bm\theta}^{\ML}}|_{\hat{\bm\theta}^{\ML}=\bm\theta_{0}}](\mathbf{\Phi}^{\top}\mathbf{\Phi})^{-1}+o(1/N^2),\\
&\Var(\hat{\bm\theta}^{\Biased}-\hat{\bm\theta}^{\ML})=o(1/N^2),
\end{align*}
where we use $\E(\hat{\bm\theta}^{\ML})=\bm\theta_{0}$ and $\Var(\hat{\bm\theta}^{\ML})=\sigma^2(\mathbf{\Phi}^{\top}\mathbf{\Phi})^{-1}$. With \eqref{eq: limit of regression matrix} and \eqref{eq: limits of b_N and its derivative}, we can obtain $\lim_{N\to\infty}N\bm{\Upsilon}_{\text{Bias}}=\bm{b}_{\star}(\bm\theta_{0})$, $\lim_{N\to\infty}N^2\bm{\Upsilon}_{\text{Var}}=\sigma^2\mathbf{b}_{\star}^{'}(\bm\theta_{0})$, and $\lim_{N\to\infty}N^2\mathbf{\Upsilon}_{\text{HOT}}=0$.
Note that $\XBias(\hat{\bm\theta}^{\Biased})$ and $\XVar(\hat{\bm\theta}^{\Biased})$ equal the limits of $N\bm{\Upsilon}_{\text{Bias}}$ and $N^2\bm{\Upsilon}_{\text{Var}}$, respectively. It leads to \eqref{eq: XMSE of biased estimator}.



\subsection{Proof of Theorem \ref{thm: explicit form of biased term}}

Note that $\log(\pi(\bm\theta))=(2-n)\log(\|\bm\theta\|_{2})+2\log(C_{1}\|\bm\theta\|_{2}+C_{2}\|\bm\theta\|_{2}^{-1})$. We can use \eqref{eq: limit of regression matrix} and \eqref{eq: explicit form of bias term} to prove that $\bm{b}_{\star}(\bm\theta_{0})$ and $\mathbf{b}_{\star}^{'}(\bm\theta_{0})$ equal the first-order and second-order derivatives of $\sigma^2 \log(\pi(\bm\theta))$ with respect to $\bm\theta$ given $\bm\theta=\bm\theta_{0}$, respectively. Then, we recall Section \ref{subsec: Proof of XMSE of Bayes estimator} and complete the proof.

\subsection{Proof of Theorem \ref{thm: generalization of conditions}}

With \eqref{eq: limit of regression matrix} replaced by $\lim_{N\to\infty}{\mathbf{\Phi}^{\top}\mathbf{\Phi}}/{N}=\sigma_{u}^2\mathbf{I}_{n}$, we can derive from \cite[Theorem 2]{JWH25} that \eqref{eq: XMSE of thetaR eta_EB} and \eqref{eq: XMSE of a generalized Bayes estimator} need to be multiplied by $1/(\sigma_{u}^2)^2$. Then it follows that Theorem \ref{thm: explicit weighting function} remains true. Similarly, $\XVar_{\#}(\hat{\bm\theta}^{\Biased})$ in Theorem \ref{thm: XMSE of biased estimator} needs to be multiplied by $1/\sigma_{u}^2$, and Theorem \ref{thm: explicit form of biased term} still holds.


\def\thesectiondis{\thesection.}                   
\def\thesubsectiondis{\thesection.\arabic{subsection}.}          
\def\thesubsubsectiondis{\thesubsection.\arabic{subsubsection}.}

\setcounter{subsection}{0}

\renewcommand{\thesection}{B}
\setcounter{lemma}{0}
\renewcommand{\thelemma}{B.\arabic{lemma}}

\renewcommand{\theequation}{B.\arabic{equation}}
\setcounter{equation}{0}

\renewcommand{\thesubsection}{\thesection.\arabic{subsection}}

\section*{Appendix B}\label{sec:Appendix B}

\begin{lemma}\label{lemma: second-order Euler equation}\cite[(2.1.2.123)]{ZP02}
For $x\in\R$, if $w(x)\in\R$ satisfies the following second-order Euler equation,
\begin{align}\label{eq: second-order Euler equation}
x^2w^{''}_{x}+axw^{'}_{x}+bw(x)=0,
\end{align}
where $w^{'}_{x}$ and $w^{''}_{x}$ denote the first-order and second-order derivatives of $w(x)$ with respect to $x$, respectively, and $a,b\in\R$ are constants independent of $x$, then $w(x)$ admits a explicit solution. When $(1-a)^2>4b$, we have
\begin{gather*}
w(x)=|x|^{(1-a)/2}(C_{1}|x|^{\mu}+C_{2}|x|^{-\mu}),\\
\text{with}\ \mu=\sqrt{|(1-a)^2-4b|}/2,\ \text{and}\ \text{arbitrary}\ C_{1},C_{2}\in\R.
\end{gather*}
\end{lemma}

\bibliographystyle{abbrv}
\bibliography{database}

\end{document}